\pdfoutput=1

\documentclass[11pt]{article}

\usepackage[]{ACL2023}

\usepackage{times}
\usepackage{latexsym}

\usepackage[T1]{fontenc}

\usepackage[utf8]{inputenc}

\usepackage{microtype}

\usepackage{inconsolata}

\usepackage{array}
\usepackage{amsmath, bm, amssymb}
\usepackage{graphicx}

\title{Characterizing and Measuring Linguistic Dataset Drift}

\author{Tyler A. Chang$^{1}$\thanks{\phantom{\_\_}Work done during an internship at AWS AI Labs.} \quad Kishaloy Halder$^2$\thanks{\phantom{\_\_}Corresponding author.} \quad Neha Anna John$^2$ \quad Yogarshi Vyas$^2$ \\
\textbf{Yassine Benajiba$^2$ \quad Miguel Ballesteros$^2$ \quad Dan Roth$^2$} \\
$^1$University of California San Diego \\
$^2$AWS AI Labs \\
\texttt{tachang@ucsd.edu} \\
\texttt{$\{$kishaloh,nehajohn,yogarshi,benajiy,ballemig,drot$\}$@amazon.com}
}

\begin{document}
\maketitle
\begin{abstract}
NLP models often degrade in performance when real world data distributions differ markedly from training data.
However, existing dataset drift metrics in NLP have generally not considered specific dimensions of linguistic drift that affect model performance, and they have not been validated in their ability to predict model performance at the individual example level, where such metrics are often used in practice.
In this paper, we propose three dimensions of linguistic dataset drift: vocabulary, structural, and semantic drift.
These dimensions correspond to content word frequency divergences, syntactic divergences, and meaning changes not captured by word frequencies (e.g. lexical semantic change).
We propose interpretable metrics for all three drift dimensions, and we modify past performance prediction methods to predict model performance at both the example and dataset level for English sentiment classification and natural language inference.
We find that our drift metrics are more effective than previous metrics at predicting out-of-domain model accuracies (mean 16.8\% root mean square error decrease), particularly when compared to popular fine-tuned embedding distances (mean 47.7\% error decrease).
Fine-tuned embedding distances are much more effective at ranking individual examples by expected performance, but decomposing into vocabulary, structural, and semantic drift produces the best example rankings of all considered model-agnostic drift metrics (mean 6.7\% ROC AUC increase).
\end{abstract}

\section{Introduction}
Dataset drift, when test data distributions differ from a model's training data, can have detrimental effects on NLP model performance \citep{broscheit-etal-2022-distributionally,do-etal-2021-predicting,koh-etal-2021-wilds}.
In real world scenarios, models are regularly monitored for potential performance degradations by comparing incoming test data with the training data \citep{elango-etal-2022-detect,nigenda-etal-2022-amazon}.
For these scenarios, researchers have proposed a variety of linguistic dataset drift metrics that aim to predict NLP model performance degradations between training and test domains
\citep{elsahar-galle-2019-annotate,ramesh-kashyap-etal-2021-domain}.

However, previous drift metrics and performance predictions suffer from several limitations.
First, previous metrics have generally been designed as holistic measures of linguistic dataset drift, despite the fact that different NLP tasks and models might be sensitive to different dimensions of linguistic drift.
Second, previous research has focused on drift metrics at the dataset level rather than the individual example level.
Not only does this require multiple labeled evaluation domain datasets to make out-of-domain performance predictions (regressions require multiple dataset-level drift values to fit to; \citealp{elsahar-galle-2019-annotate,ramesh-kashyap-etal-2021-domain}), but drift metrics are often used in practice to predict model performance when individual real world examples are streamed in real time \citep{elango-etal-2022-detect}.
We seek to overcome both of these limitations by proposing and evaluating specific dimensions of linguistic drift, predicting out-of-domain model performance at both the individual example level and the dataset level.

Specifically, we propose three dimensions of linguistic dataset drift along with corresponding drift metrics: vocabulary, structural, and semantic drift.
Because these dimensions capture distinct features that could have different effects on the performance of an NLP model, we hypothesize that decomposing into these three dimensions will allow NLP performance prediction models to better predict model performance on novel data. 
Indeed, when compared to previous model-agnostic drift metrics predicting performance on English sentiment classification and natural language inference (NLI), our metrics produce both improved predictions of dataset-level accuracies and improved rankings of individual examples by expected performance, for both in-domain and out-of-domain data (mean 16.8\% accuracy root mean square error decrease, mean 6.7\% ROC area under the curve increase).
Although we find that previously-proposed fine-tuned embedding distances \citep{elango-etal-2022-detect} are far more effective at ranking individual examples by expected performance, those distances are extremely ineffective at predicting actual model accuracies.
We conclude that decomposing linguistic drift into vocabulary, structural, and semantic drift is an effective approach for predicting out-of-domain model accuracy, and for ranking individual examples when model-agnostic metrics are desired.

\section{Related Work}
\label{sec:related-work}
Past work has quantified the drift between NLP datasets using distances between token frequency distributions or TF-IDF vectors \citep{back-2019-domain,ramesh-kashyap-etal-2021-domain,sato-etal-2022-reevaluating}, language model embedding distances \citep{feldhans-etal-2021-drift,yamshchikov-etal-2021-style}, or the ability of domain classifiers to discriminate between datasets \citep{dredze-etal-2010-kansas,elsahar-galle-2019-annotate,ruder-etal-2017-data}.
Notably, \citet{ramesh-kashyap-etal-2021-domain} find that these metrics can predict performance degradations when an NLP model is transferred from a training dataset $D_{\text{train}}$ to an out-of-domain evaluation dataset $D_{\text{eval}}$.

However, existing metrics have generally been designed as holistic measures of linguistic drift, failing to capture specific dimensions that might affect NLP model performance in different ways.
Furthermore, the traditional setup for evaluating drift metrics \citep{elsahar-galle-2019-annotate,ramesh-kashyap-etal-2021-domain} only allows for dataset-level drift metrics that predict overall model accuracy on out-of-domain datasets.
In practice, when real world examples are streamed during test time, it is desirable to predict model performance for individual examples using example-level drift metrics (i.e. drift between an example $x$ and a training dataset $D_{\text{train}}$; \citealp{elango-etal-2022-detect,nigenda-etal-2022-amazon}).
In our work, we modify the setup from \citet{ramesh-kashyap-etal-2021-domain} to predict performance for individual examples (Section \ref{sec:experiments}), using logistic regressions fitted to example-level drift metrics.
In contrast to \citet{ramesh-kashyap-etal-2021-domain}, we can fit our regressions to predict out-of-domain performance even when only a single in-domain evaluation dataset is available.

\section{Dimensions of Linguistic Drift}
\label{sec:dimensions-of-drift}
As described above, previous measures of dataset drift in NLP suffer from (1) lack of specificity and (2) lack of validation at the example level, where such metrics are often used in practice.
First, we address the lack of specificity by proposing three dimensions of linguistic dataset drift: vocabulary, structural, and semantic drift.
As in previous work, we primarily focus on \textit{domain} drift, \textit{i.e.} divergence in the input probabilities $P(x)$ rather than the joint probabilities over inputs and labels $P(x,y)$.
For each of our proposed drift dimensions, we propose a metric that quantifies the drift between an evaluation example $x$ and a training dataset $D_{\text{train}}$, allowing us to use our metrics to predict example-level model performance.
We evaluate our metrics empirically in Section \ref{sec:experiments}.

\subsection{Vocabulary Drift}
\label{sec:vocabulary-drift}
We define vocabulary drift as the divergence between content word frequencies in two text samples.
Content words are defined as open class words that generally contain substantial semantic content (e.g. nouns, verbs, adjectives, and adverbs), contrasted with function words that primarily convey grammatical relationships (e.g. prepositions, conjunctions, and pronouns; \citealp{bell-etal-2009-predictability,segalowitz-lane-2000-lexical}).
By restricting our vocabulary drift definition to content word distributions, we capture vocabulary differences between two text datasets without the confounds of structural features.
For example, \textit{``The student ate the sandwich''} and \textit{``A sandwich was eaten by a student''} would have low vocabulary drift after excluding function words.
Notably, our definition of vocabulary drift is designed to include drift in word choice, regardless of the semantic similarity between chosen words; for example, \textit{``The dog was happy''} and \textit{``The beagle was ecstatic''} would have high vocabulary drift due to differing word choice, despite their high semantic similarity.
This property is useful because NLP models are often sensitive to changes in word choice even if datasets are semantically similar \citep{hu-etal-2019-shot,misra-etal-2020-exploring}.

Formally, to quantify the vocabulary drift between an evaluation example $x$ and a training dataset $D_{\text{train}}$, we compute the cross-entropy between content word frequencies in $x$ and $D_{\text{train}}$ as:
\begin{equation}
\frac{1}{|x_{\text{content}}|} \sum_{w \in x_{\text{content}}} \text{log}(P_{\text{train\_content}}(w)).
\end{equation}
Here, $x_{\text{content}}$ is the set of content words in example $x$, and $P_\text{train\_content}(w)$ is the frequency (restricted to content words) of word $w$ in the training dataset.
Our vocabulary drift metric is equal to the log-perplexity (training loss) of a unigram language model restricted to content words, trained on $D_{\text{train}}$ and evaluated on $x$.
We annotate content words using the {\tt spaCy} tokenizer and part-of-speech (POS) tagger \citep{honnibal-etal-spacy}, defining content words as those with an open class Universal POS tag (nouns, verbs, adjectives, adverbs, and interjections; \citealp{nivre-etal-2020-universal}) and excluding stop words in {\tt spaCy}.

\subsection{Structural Drift}
\label{sec:structural-drift}
In contrast to vocabulary drift, structural drift captures divergences between the syntactic structures in two text samples.
For example, \textit{``Yesterday, I was surprised by a dog''} and \textit{``Usually, she is recognized by the audience''} would have low structural drift despite high vocabulary drift.
Previous work in discourse analysis has attempted to quantify structural similarity separately from semantic similarity in natural conversations, although their metrics are not directly applicable to NLP datasets due to computational limitations \citep{boghrati-etal-2018-conversation}.\footnote{The CASSIM structural similarity metric in \citet{boghrati-etal-2018-conversation} is based on tree-edit distances between all sentence pairs, which is slow to compute even for relatively small NLP datasets.}
Structural drift has also been studied in machine translation, primarily considering structural divergence between parallel text in different languages \citep{dave-etal-2004-interlingua,deng-xue-2017-translation,dorr-1990-solving,saboor-khan-2010-lexical}; in our work, we focus on divergences between non-parallel monolingual text.

We quantify the structural drift between an example $x$ and $D_{\text{train}}$ using the cross-entropy between the true POS tag sequence for $x$ and the predictions of a POS 5-gram model trained on POS tag sequences in $D_{\text{train}}$.
This metric captures the divergence between syntactic structures in $x$ and $D_{\text{train}}$ using 5-gram sequences, abstracting away from semantic content and vocabulary by considering only the POS tag for each word \citep{axelrod-etal-2015-class,nerbonne-wiersma-2006-measure}.
Formally, we compute:
\begin{equation}
\frac{1}{|x|} \sum_{i=1}^{|x|} \text{log}(P_{\text{train}}(\text{tag}_i | \text{tag}_{i-1}, ..., \text{tag}_{i-4})).
\end{equation}
We pad the beginning of the POS tag sequence with [SEP] tokens, and we only annotate examples with structural drift if they contain at least two non-[SEP] tokens.
As with our vocabulary drift metric, we annotate POS tags using the {\tt spaCy} tokenizer and POS tagger. 

\subsection{Semantic Drift}
\label{sec:semantic-drift}
Finally, we consider semantic drift, defined as any divergence in semantic meaning between two text samples.
Semantic drift is closely related to both vocabulary and structural drift; the words and syntactic structures used in a sentence are closely tied to the meaning of that sentence, particularly under compositional assumptions of language \citep{szabo-2022-compositionality}.
However, there are notable cases where semantic drift is independent from vocabulary and structural drift.
For example, \textit{``I saw the doctor''} and \textit{``I took a trip to the hospital''} have high vocabulary and structural drift under our definitions, despite similar semantic meaning.
Conversely, some sentences have different meanings or connotations across time and contexts, despite remaining identical in both vocabulary and structure (e.g. the word \textit{``sick''} in \textit{``That salamander is sick!''} can mean very cool or physically ill depending on the context).

Many of these semantic similarities and differences can be quantified using contextualized embeddings from modern language models \citep{briakou-carpuat-2020-detecting,devlin-etal-2019-bert,liu-etal-2020-survey,sun-etal-2022-sentence}, which we include in our drift metric experiments (Section \ref{sec:experiments}).
However, when identifying individual dimensions of linguistic drift, we seek to identify dimensions that are both interpretable and relatively independent from one another, to better isolate specific dimensions that impact NLP model performance.
Language model embeddings reflect vocabulary and structural properties of sentences as well as semantic properties \citep{hewitt-manning-2019-structural,tenney-etal-2019-learn}, and thus they are less effective for pinpointing interpretable effects that are specific to semantic drift.

\paragraph{Lexical Semantic Change.}
Instead, we consider lexical semantic change, in which a word's meaning changes between two datasets while its surface form remains the same \citep{gulordava-baroni-2011-distributional,kulkarni-etal-2015-statistically,sagi-etal-2009-semantic,tahmasebi-etal-2021-survey}.
Past work has quantified a token's lexical semantic change $\text{LSC}_{D_1 \leftrightarrow D_2}(w)$ using the mean pairwise cosine distance between contextualized RoBERTa embeddings for that token in two different datasets $D_1$ and $D_2$ \citep{giulianelli-etal-2020-analysing,laicher-etal-2021-explaining}.
Motivated by this metric, we quantify the lexical semantic change between an evaluation example $x$ and a training dataset $D_{\text{train}}$ using the mean lexical semantic change between $x$ and $D_{\text{train}}$ for all content tokens $w$ shared between $x$ and $D_{\text{train}}$:
\begin{equation}
\frac{1}{|x_{\text{content}}|} \sum_{w \in x_{\text{content}}} \text{LSC}_{x \leftrightarrow D_{\text{train}}}(w).
\end{equation}
Here, $\text{LSC}_{x \leftrightarrow D_{\text{train}}}(w)$ is the mean pairwise cosine distance between embeddings for $w$ in example $x$ and dataset $D_{\text{train}}$, using a non-fine-tuned RoBERTa model.
Again, we define content tokens as tokens that are annotated with an open class POS tag anywhere in the Universal Dependencies English dataset, excluding stop words \citep{nivre-etal-2020-universal}.\footnote{
We exclude non-content tokens for lexical semantic change because non-content token embeddings (e.g. for function words and punctuation) are more likely to encode structural drift information rather than lexical semantic change.
Contextualized token embeddings are computed as the mean of the token representations in the last two RoBERTa layers before fine-tuning \citep{elango-etal-2022-detect}.
}
While this lexical semantic change metric is still based on contextualized embeddings, matching embeddings based on token surface forms allows us to minimize effects of vocabulary and structural drift, as compared to matching each example representation with all other example representations regardless of surface form.
Of course, lexical semantic change is just one type of semantic drift; future work might consider other types of semantic drift that are independent from vocabulary and structural drift.


\section{Experiments}
\label{sec:experiments}
Previous work has evaluated drift metrics by assessing their ability to predict out-of-domain model performance at the dataset-level using dataset-level metrics (e.g. \citealp{ramesh-kashyap-etal-2021-domain}; Section \ref{sec:related-work}).
We extend this work by predicting individual example-level performance (probabilities of getting individual examples correct) along with dataset-level accuracies, using drift metrics between each example $x$ and the training dataset $D_{\text{train}}$.
Using these example-level metrics instead of dataset-level metrics allows us to fit regressions predicting model performance using only a set of examples (e.g. using only the in-domain evaluation set), rather than a set of multiple evaluation datasets covering different domains.
Thus, our approach can be used in common real world scenarios where labeled data is available only in one domain.
In our experiments, we compare previous drift metrics with our proposed metrics for vocabulary, structural, and semantic drift, evaluating whether decomposing linguistic drift into these three dimensions improves NLP model performance predictions.
\footnote{Code is available at \\ \url{https://github.com/amazon-science/characterizing-measuring-linguistic-drift}.}

\subsection{Datasets}
\label{sec:datasets}
We evaluate cross-domain transfer performance for language models fine-tuned on sentiment classification (split by product category or review year) and natural language inference (NLI, split by source domain).
Because these tasks output one prediction per sequence, they allow us to directly evaluate sequence-level (i.e. example-level) drift metrics.

\paragraph{Amazon Reviews (product categories).}
For sentiment classification, we consider the Amazon reviews dataset, containing customer-written product reviews for 43 different product categories \citep{amazon-2017-customer}.
As in \citet{blitzer-etal-2007-biographies}, we label 1- and 2-star reviews as negative, and 4- and 5-star reviews as positive.
We sample up to 100K polarity-balanced reviews from each product category, considering each category as a domain.
For each product category, we use a 70/20/10\% split for training, evaluation and test datasets.

\paragraph{Amazon Reviews (temporal split).}
Next, we consider the same Amazon reviews dataset for sentiment classification, but we define domains by review date rather than by product category.
We generate a category-balanced and polarity-balanced sample for each year between 2001 and 2015 (inclusive) by sampling up to 5K polarity-balanced reviews from each product category for each year, sampling the same number of reviews each year for any given category. 
The resulting dataset has 33K training examples, 5K evaluation examples, and 5K test examples for each year, similar to \citet{agarwal-nenkova-2022-temporal}, but balanced for product category and polarity.

\paragraph{MultiNLI.}
Finally, we consider the MNLI dataset for natural language inference (NLI), covering five training domains and ten evaluation domains, including government documents, pop culture articles, and transcribed telephone conversations \citep{williams-etal-2018-broad}.
Each training domain has approximately 77K training examples, and each evaluation domain has approximately 2K evaluation examples.

\subsection{Models}
\label{sec:models}
We fine-tune a RoBERTa base-size model $\mathcal{M}$ for each training domain for each task, using batch size 32, learning rate 2e-5, and four epochs through the training data \citep{liu-etal-2019-roberta}.
Because there are only five training domains for MNLI, we run five fine-tuning runs per MNLI training domain.
Full fine-tuning details and hyperparameters are listed in Appendix \ref{app:finetuning-details}.
We evaluate each model on each evaluation domain; to simulate realistic scenarios for temporal data, we evaluate only on future years for models trained on temporal splits.

\subsection{Drift metrics}
\label{sec:drift-metrics}
We consider drift metrics between individual evaluation examples $x$ and training datasets $D_{\text{train}}$.
First, we consider our vocabulary, structural, and semantic drift metrics from Section \ref{sec:dimensions-of-drift}.
Initial motivations and theoretical examples of how these three dimensions differ are described in Section \ref{sec:dimensions-of-drift}, but the dimensions are not perfectly independent. Empirically, Pearson correlations between our vocabulary, structural, and semantic drift metrics range from $0.10$ to $0.50$ across the different tasks.
For comparison, we also consider drift metrics from past work: token frequency divergences and embedding cosine distances.
With the exception of the fine-tuned embedding distances, all of our metrics are model-agnostic,
meaning they are not dependent on the internals of the fine-tuned model. 

\paragraph{Token frequency divergences.}
We compute the Jensen-Shannon (JS) divergence between the token frequency distribution for each example $x$ and each training dataset $D_{\text{train}}$.
This divergence has been shown to correlate with out-of-domain model performance when computed at the dataset-level (i.e. between an entire evaluation set $D_{\text{eval}}$ and the training set $D_{\text{train}}$; \citealp{ramesh-kashyap-etal-2021-domain}), and it has been recommended as a metric for training dataset selection \citep{ruder-etal-2017-data}. 

However, because example-level token frequency distributions are quite sparse \citep{ruder-etal-2017-data}, we also consider the cross-entropy between each example frequency distribution and each training frequency distribution (i.e. the loss of a unigram language model).
The resulting token frequency cross-entropy is equivalent to our vocabulary drift metric, but using the RoBERTa tokenizer and without the restriction to content words.

\paragraph{Embedding cosine distances.}
We compute embeddings for training and evaluation examples $x$ by taking the mean over all tokens in $x$ and the last two RoBERTa layers, either before or after task fine-tuning (i.e. pre-trained or fine-tuned; \citealp{elango-etal-2022-detect}).
We note that the pre-trained RoBERTa model is still the same model that is fine-tuned for each task, potentially leading to overly optimistic results for the pre-trained embedding cosine distances; this caveat also holds for our semantic drift metric, which relies on pre-trained embeddings.
For the embedding cosine distance drift metrics, we compute the mean cosine distance between the embedding for evaluation example $x$ and each example in the training dataset $D_{\text{train}}$ \citep{nigenda-etal-2022-amazon}.\footnote{For efficiency, we compute mean pairwise cosine distances using the method described in Appendix \ref{app:efficient-cosine-distances}.}

\setlength{\belowcaptionskip}{-0.3cm}
\begin{figure}
    \centering
    \includegraphics[height=4.0cm]{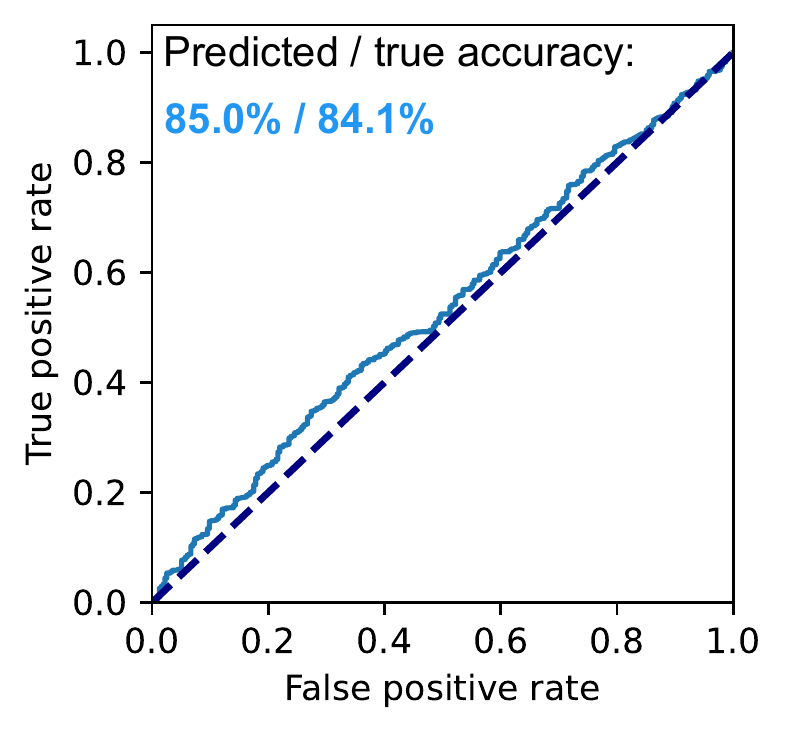}
    \hspace{-0.45cm}
    \includegraphics[height=4.0cm]{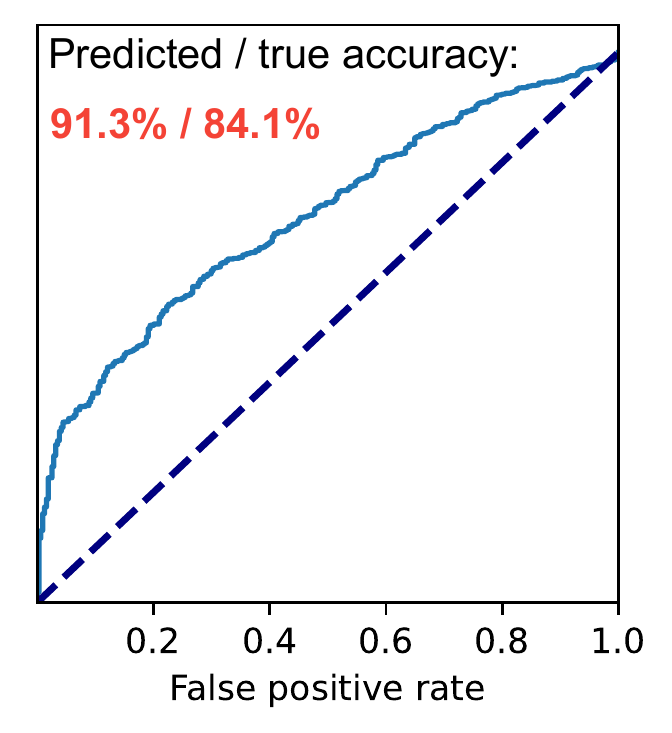}
    \caption{ROC curves predicting whether a model will get individual examples correct based on vocabulary, structural, and semantic drift (left) vs. fine-tuned embedding distances (right), for an MNLI model transferred from the telephone to fiction domain. The true model accuracy degradation is $87.9\% \to 84.1\%$. Here, our decomposed drift metrics produce worse example rankings than the fine-tuned embedding distances (ROC AUC $0.532$ vs. $0.702$), but a much better dataset-level accuracy prediction (absolute error $0.9\%$ vs. $7.2\%$). We observe this pattern to hold across domains and datasets (Table \ref{tab:main-results}). Still, our decomposed metrics outperform previous model-agnostic metrics by all evaluation criteria.}
    \label{fig:roc-example}
\end{figure}
\setlength{\belowcaptionskip}{0cm}

\subsection{Predicting Model Performance}
For each drift metric (or set of drift metrics) and each model $\mathcal{M}$ trained on dataset $D_{\text{train}}$, we fit a logistic regression predicting whether $\mathcal{M}$ will get example $x$ correct (i.e. a ``positive'' example), based on the drift metric(s) between $x$ and $D_{\text{train}}$.
The regression input is the considered drift metric(s) from $x$ to $D_{\text{train}}$, and the label is 1 if $\mathcal{M}$ predicts $x$ correctly, and 0 otherwise.\footnote{In cases where we input multiple drift metrics into the logistic regression, we exclude interaction terms; interaction terms generally resulted in worse out-of-domain performance predictions, based on both ROC AUCs and RMSEs.}
We fit the logistic regression for all $x$ in the in-domain evaluation dataset, mimicking a scenario where labeled evaluation data is only available in the same domain as training.
This allows us to test whether regressions fitted only to in-domain examples can extrapolate to out-of-domain examples.

We evaluate the logistic regressions on both in-domain and out-of-domain evaluation examples.
Each regression produces a predicted probability of ``positive'' (getting an example correct) for each example.\footnote{For in-domain evaluation example predictions, we use 5-fold cross-validation, fitting regressions to only 80\% of the in-domain evaluation dataset per fold.}
For dataset-level accuracy predictions, we compute the mean predicted ``positive'' probability over all examples in each evaluation dataset $D_{\text{eval}}$, equal to the expected value of model accuracy on $D_{\text{eval}}$ based on the example-level probabilities.


\setlength{\belowcaptionskip}{-0.3cm}
\setlength\tabcolsep{3pt}
\begin{table*}[t]
    \centering
    \small
    \renewcommand{\arraystretch}{1.3}
    \begin{tabular}{|p{4cm}|
    >\raggedleft p{1.3cm} >\raggedleft p{0.9cm} >\raggedleft p{1.0cm}|
    >\raggedleft p{1.3cm} >\raggedleft p{0.9cm} >\raggedleft p{1.0cm}|
    >\raggedleft p{1.3cm} >\raggedleft p{0.9cm}
    >{\raggedleft\arraybackslash} p{1.0cm}|}
        \cline{2-10}
        \multicolumn{1}{c|}{} & \multicolumn{3}{c|}{\textbf{Sentiment (categories)}} & \multicolumn{3}{c|}{\textbf{Sentiment (temporal)}} & \multicolumn{3}{c|}{\textbf{MNLI (source domains)}} \\
        \multicolumn{1}{c|}{} & \multicolumn{1}{c}{In-domain} & \multicolumn{2}{c|}{Out-of-domain} & \multicolumn{1}{c}{In-domain} & \multicolumn{2}{c|}{Out-of-domain} & \multicolumn{1}{c}{In-domain} & \multicolumn{2}{c|}{Out-of-domain} \\
        \hline
        \textbf{Drift metric(s)} & \textbf{ROC \\ AUC $\uparrow$} & \textbf{ROC AUC $\uparrow$} & \textbf{RMSE \% $\downarrow$} & \textbf{ROC \\ AUC $\uparrow$} & \textbf{ROC AUC $\uparrow$} & \textbf{RMSE \% $\downarrow$} & \textbf{ROC \\ AUC $\uparrow$} & \textbf{ROC AUC $\uparrow$} & \textbf{RMSE \% $\downarrow$} \\
        \hline
        Baseline (no-performance drop) & 0.500 & 0.500 & 100.0\% & 0.500 & 0.500 & 100.0\% & 0.500 & 0.500 & 100.0\% \\
        Token frequency JS-div & 0.512 & 0.517 & 98.4\% & 0.519 & 0.528 & 106.2\% & 0.496 & 0.503 & 118.8\% \\
        \phantom{\_} (\citealp{ramesh-kashyap-etal-2021-domain}; & & & & & & & & & \\
        \phantom{\_} \citealp{ruder-etal-2017-data}) & & & & & & & & & \\
        Token frequency cross-entropy & 0.540 & 0.551 & 71.4\% & 0.543 & 0.557 & 97.3\% & 0.500 & 0.512 & 96.8\% \\
        Cosine distance (pre-trained) & 0.535 & 0.558 & 93.6\% & 0.534 & 0.559 & 91.8\% & 0.484 & 0.508 & 107.5\% \\
        \phantom{\_} (\citealp{ramesh-kashyap-etal-2021-domain}) & & & & & & & & & \\
        Combined prev. model-agnostic & 0.551 & 0.557 & 70.3\% & 0.554 & 0.562 & 142.1\% & 0.520 & 0.514 & 99.8\% \\
        \hline
        Vocabulary drift & 0.561 & 0.570 & \textbf{51.8\%} & 0.552 & 0.571 & 105.8\% & 0.474 & 0.500 & 81.5\% \\
        Structural drift & 0.572 & 0.575 & 91.4\% & 0.568 & 0.581 & 146.1\% & 0.516 & \textbf{0.531} & 80.6\% \\
        Semantic drift & 0.586 & 0.591 & 58.4\% & 0.565 & 0.586 & 110.4\% & 0.516 & 0.521 & \textbf{79.1\%} \\
        Vocabulary, structural, \newline \phantom{\_\_} semantic drift & \textbf{0.597} & \textbf{0.601} & 52.4\% & \textbf{0.578} & \textbf{0.596} & \textbf{84.8\%} & \textbf{0.525} & \textbf{0.531} & 81.0\% \\
        \hline
        \multicolumn{10}{c}{} \\
        \hline
        Model-dependent: & & & & & & & & & \\
        Cosine distance (fine-tuned) & \textbf{0.845} & \textbf{0.822} & 81.9\% & \textbf{0.852} & \textbf{0.834} & 236.7\% & \textbf{0.699} & \textbf{0.683} & 141.9\% \\
        \phantom{\_} \citep{nigenda-etal-2022-amazon} & & & & & & & & & \\
        \hline
    \end{tabular}
    \caption{Mean ROC AUCs and RMSEs using different drift metrics to predict model performance, comparing our metrics (vocabulary, structural, and semantic drift) with previous metrics.
    ROC AUCs indicate the quality of example rankings by expected performance, and RMSEs (as percentages of the baseline error) indicate the quality of the actual accuracy predictions.
    Given in-domain accuracy $p$, the baseline predicts out-of-domain accuracy $p$ and an equal probability $p$ of getting any individual example correct. All metrics are model-agnostic except the fine-tuned embedding cosine distances.}
    \label{tab:main-results}
\end{table*}
\setlength{\belowcaptionskip}{0cm}

\subsection{Evaluating Performance Predictions}
\label{sec:evaluating-predictions}
We use ROC curves to evaluate example-level performance predictions, both in-domain and out-of-domain, and we use root mean square errors (RMSEs) to evaluate out-of-domain dataset-level accuracy predictions.

\paragraph{ROC AUC.} For each logistic regression, predicting positive examples (correct model predictions) from a given drift metric and for a given model, we compute the area under the ROC curve for in-domain and out-of-domain examples.
An ROC curve plots recall (proportion of true positives identified) over the false positive rate for different probability thresholds.
In our case, a higher ROC AUC indicates that the input drift metric can generally predict more true positives (examples the model gets correct) for a given false positive rate.
However, ROC curves are dependent only on the rankings of examples by predicted positive probabilities \citep{tang-etal-2010-compare}; the raw probabilities of correct model predictions do not affect the ROC AUC as long as the example ranking is preserved.
From this perspective, a higher ROC AUC indicates that evaluation examples are ranked roughly in order of expected performance; examples with higher predicted probabilities are more likely to be predicted correctly by the model.
For each drift metric, we compute the mean ROC AUC over all trained models $\mathcal{M}$, for in-domain and out-of-domain examples.

\paragraph{RMSE.}
Because ROC AUCs depend only on the ranking of evaluation examples, they do not capture whether the predicted positive probabilities (probabilities of correct predictions) are actually reflective of model accuracies.
For example, a given drift metric can achieve a high ROC AUC by ranking evaluation examples accurately, even if the mean probability (expected model accuracy) is far from the true model accuracy for $D_{\text{eval}}$ (e.g. Figure \ref{fig:roc-example}).

Thus, for each drift metric, we also compute the RMSE comparing expected model accuracy (mean positive probability over all examples in $D_{\text{eval}}$) to actual model accuracy on $D_{\text{eval}}$.
We compute RMSEs over all models $\mathcal{M}$ and their corresponding out-of-domain datasets $D_{\text{eval}}$.
We report RMSEs as percentages of a baseline RMSE that predicts out-of-domain accuracy on $D_{\text{eval}}$ to be the same as the in-domain evaluation accuracy (i.e. predicting no out-of-domain performance drop).
Our reported RMSE percentages indicate the percentage of accuracy prediction error that remains when using a given drift metric, relative to the baseline.

To summarize, we compute the predicted accuracy RMSE and the mean ROC AUC for each drift metric and for each task.
ROC AUC measures how well a drift metric ranks the evaluation examples (examples with higher ``positive'' probabilities should be more likely to be predicted correctly by the model), while RMSE measures how well the drift metric predicts actual model accuracy (mean probabilities should be close to the true model accuracy).
An ideal drift metric should have high ROC AUC and low RMSE.

\section{Results}
\label{sec:results}
The mean accuracy change ($\pm$standard deviation; in raw accuracy percentage difference) from in-domain to out-of-domain evaluation is $-1.04\pm1.02\%$ for sentiment classification across product categories, $-0.20\pm0.57\%$ for sentiment classification across years, and $-1.83\pm2.91\%$ for MNLI across source domains.
Notably, in many cases, accuracy improves for out-of-domain evaluation (e.g. MNLI fiction $\to$ government).
Results predicting out-of-domain evaluation accuracies (RMSEs) and example-level performance (ROC AUCs) from different drift metrics are reported in Table \ref{tab:main-results}.

\subsection{Ranking Examples (ROC AUC)}
Mean ROC AUCs for different drift metrics are reported in Table \ref{tab:main-results}, for both in-domain and out-of-domain evaluation examples.
Recall that a higher ROC AUC indicates that higher scoring examples (as ranked by the logistic regression) are more likely to be predicted correctly by the model.

\paragraph{Decomposing drift improves rankings.} Using vocabulary, structural, and semantic drift as input features into the logistic regressions results in higher ROC AUCs than any of the previous model-agnostic drift metrics, for all three multi-domain datasets and for both in-domain and out-of-domain examples (top section of Table \ref{tab:main-results}).
Across the three datasets, this decomposed drift improves ROC AUCs by an average of $0.039$ for in-domain examples and $0.033$ for out-of-domain examples when compared to the best model-agnostic drift metric from previous work.

To ensure that this is not simply a result of including three different metrics in the regression, we also consider the combination of all three model-agnostic metrics from previous work (``combined previous model-agnostic'' in Table \ref{tab:main-results}: token frequency JS-divergence, token frequency cross-entropy, and pre-trained embedding cosine distance).
For all three datasets, the combination of previous metrics still results in worse ROC AUCs than the combination of vocabulary, structural, and semantic drift, for both in-domain and out-of-domain examples.
This indicates that decomposing into vocabulary, structural, and semantic drift results in better rankings of individual examples by expected performance than previous model-agnostic drift metrics.

\paragraph{Fine-tuned embeddings lead to the best rankings.}
However, fine-tuned (model-dependent) embedding cosine distances result in by far the best rankings of examples by expected performance (higher ROC AUCs).
Indeed, this is the recommended drift metric when examples need to be ranked relative to one another or relative to some threshold (e.g. when there is some threshold drift value to flag examples; \citealp{elango-etal-2022-detect,nigenda-etal-2022-amazon}); our results validate this approach.
Notably, the fine-tuned embedding distances produce quality rankings even for out-of-domain examples, despite work suggesting that fine-tuning affects the in-domain representation space differently from the out-of-domain representation space in language models \citep{merchant-etal-2020-happens}.
Our results indicate that despite these differences between the in-domain and out-of-domain fine-tuned spaces, the fine-tuned embedding distances can still be used to rank both in-domain and out-of-domain examples by expected performance.

That said, fine-tuned embedding distances require access to the internal representations of a given model; model-agnostic metrics are still useful in cases where only model outputs can be observed, or when the same drift metric needs to apply to multiple models.
For these use cases, our decomposed vocabulary, structural, and semantic drift metrics outperform previous model-agnostic metrics.
Furthermore, as we observe in the next section, our decomposed drift metrics result in drastically better out-of-domain accuracy predictions than fine-tuned embedding distances, despite worse rankings of individual examples.

\subsection{Predicting Model Accuracy (RMSE)}
As described in Section \ref{sec:evaluating-predictions} and shown in Figure \ref{fig:roc-example}, a given drift metric can produce quality rankings of examples even if the raw predicted accuracies are far from the true model accuracies.
Thus, as reported in Table \ref{tab:main-results}, we evaluate RMSEs using different drift metrics to predict model accuracies for out-of-domain evaluation datasets.\footnote{We only consider accuracy prediction RMSEs for out-of-domain datasets because sufficiently sized in-domain datasets have very low variation in model accuracy.}

\paragraph{Decomposed drift has the best accuracy predictions.}
Decomposing into vocabulary, structural, and semantic drift results in better dataset-level accuracy predictions (lower RMSEs) than any previous drift metric(s), for all three multi-domain datasets.
Accuracy predictions based on individual dimensions vary (e.g. individual dimensions are sometimes better than including all three dimensions), but predicting out-of-domain accuracy from all three dimensions results in reliably low errors compared to previous metrics.
Across the three datasets, our decomposed drift results in an average decrease of $16.8\%$ in accuracy prediction error (RMSE) when compared to the best metric from previous work.

\paragraph{Fine-tuned embedding distances have poor accuracy predictions.}
The fine-tuned embedding distances result in worse out-of-domain accuracy predictions (higher RMSEs) than our decomposed vocabulary, structural, and semantic drift for all three multi-domain datasets.
Notably, they have by far the worst out-of-domain accuracy predictions of any drift metric for MNLI and sentiment classification split temporally.
Across all three datasets, fine-tuned embedding distances result in an average of $2.03$x more error than our decomposed vocabulary, structural, and semantic drift.
This contrasts with fine-tuned embedding distances' ability to rank individual examples by expected performance better than any other metric(s).
This suggests that despite maintaining \textit{relative} distances that are predictive of relative model performance for individual examples (high ROC AUCs), fine-tuning adjusts the example embeddings such that \textit{raw} distances are not predictive of raw out-of-domain accuracies (high RMSEs).
Concretely, the logistic regressions fit to fine-tuned embedding distances yield example-level probabilities that are highly predictive of \textit{relative} model performance between out-of-domain examples, but quite far from the \textit{actual} expected probabilities of getting each example correct.
In practice, this suggests that fine-tuned embedding distances should be used in scenarios where the relative performance of evaluation examples is important (e.g. establishing drift threshold values), but they should not be used to predict actual out-of-domain model accuracies.

\section{Discussion}
We find that decomposing linguistic dataset drift into our proposed vocabulary, structural, and semantic drift metrics leads to improved out-of-domain dataset-level accuracy predictions for sentiment classification and NLI.
Furthermore, our decomposed drift metrics produce better rankings of individual examples by expected performance than previous model-agnostic drift metrics (e.g. token frequency divergences and pre-trained embedding distances), both in-domain and out-of-domain.
Although fine-tuned embedding distances produce by far the best example rankings, they also produce 
egregiously incorrect out-of-domain model accuracy predictions.
Our results suggest that fine-tuned embedding distances should still be used in cases where examples need to be ranked by expected performance (e.g. relative to a cutoff value, as in \citealp{elango-etal-2022-detect}).
Vocabulary, structural, and semantic drift should be used in cases where either (1) the internal states of a model are unavailable, which is increasingly common as models are accessed through external APIs, (2) the same metric values need to be applied across multiple models (i.e. model-agnostic metrics), or (3) raw model accuracy predictions are desired.

Our work also opens up future directions of research studying specific effects of linguistic dataset drift on NLP model performance.
First, future work might assess whether there are systematic effects of particular drift dimensions on specific tasks or model architectures.
Second, it might consider new types of linguistic drift, potentially extending beyond domain drift (drift in $P(x)$) to consider concept drift $P(y | x)$ in NLP \citep{webb-etal-2016-characterizing}.
Finally, future work might investigate methods of quantifying drift in natural language generation, where the outputs $y$ are linguistic data.
Our work lays the groundwork for these future investigations.

\section{Conclusion}
We propose three dimensions of linguistic dataset drift---vocabulary, structural, and semantic drift---and we modify previous performance prediction methods to predict NLP model performance at the individual example level along with the dataset level.
We validate existing drift metrics for particular use cases (e.g. fine-tuned embedding distances for example ranking), and we highlight complementary use cases where our decomposed drift metrics outperform previous metrics (e.g. when predicting model accuracies or when using model-agnostic metrics).
Our work lays the foundation for future research into specific and interpretable dimensions of linguistic dataset drift, improving our ability to predict NLP model performance on real world data.

\section*{Limitations}
Our work has several limitations.
First, our experiments are limited by the multi-domain datasets available for sequence classification tasks, limiting both our task coverage (sentiment classification and NLI) and domain type coverage (product categories, temporal splits, and text source domains).
Future work can evaluate our drift metrics on token classification tasks or even sequence-to-sequence tasks by predicting sequence-level performance (e.g. proportions of correct tokens, or example-level BLEU scores; \citealp{papineni-etal-2002-bleu}) from our example-level drift metrics.
Past work has already considered dataset-level drift metrics and performance predictions for token classification tasks such as named entity recognition (NER) and part-of-speech (POS) tagging \citep{ramesh-kashyap-etal-2021-domain,rijhwani-preotiuc-pietro-2020-temporally}, and example-level drift metrics have been used in machine translation for training data example selection \citep{axelrod-etal-2011-domain,wang-etal-2017-instance}.
We hope that future work will evaluate example-level drift metrics in their ability to predict NLP model performance on this wider variety of tasks.

Second, we only consider simple logistic regressions to predict whether individual examples will be predicted correctly by different models.
More complex classifiers (e.g. XGBoost; \citealp{chen-guestrin-2016-xgboost}) might improve performance predictions, particularly if more drift metrics are included as inputs, or if raw example features are included (e.g. sequence length; \citealp{ye-etal-2021-towards}).
Our three dimensions of linguistic drift (vocabulary, structural, and semantic drift) represent just one way of decomposing linguistic dataset drift into distinct dimensions.
We hope that future work will explore novel dimensions of linguistic drift, identifying new ways of integrating different drift metrics into NLP model performance predictions across tasks and domains.


\bibliography{anthology,custom}
\bibliographystyle{acl_natbib}

\appendix

\section{Appendix}

\begin{table}[ht]
    \centering
    \small
    \renewcommand{\arraystretch}{1.2}
    \begin{tabular}{|>{\raggedright}p{3.4cm}|p{3.5cm}|}
        \hline
        \textbf{Hyperparameter} & \textbf{Value} \\
        \hline
        Learning rate decay & Linear \\
        Warmup steps & 10\% of total \\
        Learning rate & 2e-5 \\
        Adam $\epsilon$ & 1e-6 \\
        Adam $\beta_1$ & 0.9 \\
        Adam $\beta_2$ & 0.999 \\
        Attention dropout & 0.1 \\
        Dropout & 0.1 \\
        Weight decay & 0.0 \\
        Batch size & 32 \\
        Train steps & 4 epochs \\
        \hline
    \end{tabular}
    \caption{Sentiment classification and NLI fine-tuning hyperparameters for the RoBERTa-base models in Section \ref{sec:models}.}
    \label{tab:finetune-hyperparams}
\end{table}

\subsection{Model fine-tuning details}
\label{app:finetuning-details}
For each sentiment classification and NLI training domain in Section \ref{sec:experiments}, we fine-tune a RoBERTa base-size model using the hyperparameters in Table \ref{tab:finetune-hyperparams} and the pre-trained RoBERTa model from Hugging Face, containing approximately 123M parameters \cite{liu-etal-2019-roberta,wolf-etal-2020-transformers}.
Because there are only five training domains for MNLI, we run five fine-tuning runs per MNLI training domain; otherwise, we run one fine-tuning run per domain (43 domains for sentiment classification split by product category, 15 domains for sentiment classification split by review year).
All models are fine-tuned using one Tesla V100 GPU, taking about two hours per model.

\subsection{Efficient cosine distance computations}
\label{app:efficient-cosine-distances}
In Section \ref{sec:drift-metrics}, we compute the mean cosine distance between each evaluation example embedding $\bm{x}$ and all training example embeddings from $D_{\text{train}}$.
Each example embedding is computed by taking the mean over all tokens in the example and the last two RoBERTa layers (before or after fine-tuning, as specified; \citealp{elango-etal-2022-detect}).
Mean embedding cosine distances are also computed for individual tokens when quantifying lexical semantic change in Section \ref{sec:semantic-drift}.
To avoid saving the embedding for each example in $D_{\text{train}}$ and computing each cosine distance individually, we note that the mean pairwise cosine similarity between a set of vectors $U$ and $V$ is:
\begin{align*} \underset{u \in U, v \in V}{\text{Mean}} \left( \text{cos}(u, v) \right) &= \frac{1}{|U| |V|} \sum_{u \in U, v \in V} \frac{\langle u,  v \rangle}{||u|| \cdot ||v||} \end{align*}
\begin{align*}
\hspace{0.3cm} &= \frac{1}{|U| |V|} \sum_{u \in U} \sum_{v \in V} \left\langle \frac{u}{||u||}, \frac{v}{||v||} \right\rangle \\
&= \frac{1}{|U| |V|} \sum_{u \in U} \left\langle \frac{u}{||u||}, \sum_{v \in V} \frac{v}{||v||} \right\rangle \\
&= \frac{1}{|U| |V|} \left\langle  \sum_{u \in U} \frac{u}{||u||}, \sum_{v \in V} \frac{v}{||v||} \right\rangle \\
&= \left\langle \frac{1}{|U|} \sum_{u \in U} \frac{u}{||u||}, \frac{1}{|V|} \sum_{v \in V} \frac{v}{||v||} \right\rangle \\
&= \left\langle \underset{u \in U}{\text{Mean}} \left( \frac{u}{||u||} \right), \underset{v \in V}{\text{Mean}} \left( \frac{v}{||v||} \right) \right\rangle
\end{align*}
In other words, we only need to compute the dot product between the mean normed vector for $U$ and $V$.
For our uses, when computing the mean cosine distance between an example embedding $\bm{x}$ and all training example embeddings from $D_{\text{train}}$, we need only compute one minus the dot product between the normed $\bm{x}$ and the mean normed embedding over all examples in $D_{\text{train}}$.
This way, we only need to store one vector (the mean normed embedding) for the entire training set, rather than one vector per training example.

\end{document}